\documentclass[runningheads]{llncs}

\usepackage{eccv}
\usepackage{eccvabbrv}

\usepackage{multirow}
\usepackage[inline]{enumitem}
\usepackage{graphicx}
\usepackage{booktabs}

\usepackage[accsupp]{axessibility}  % Improves PDF readability for those with disabilities.

\usepackage{hyperref}
\usepackage{float}
\usepackage{orcidlink}

\begin{document}
\title{MVTN: A Multiscale Video Transformer Network for Hand Gesture Recognition} 

\titlerunning{MVTN: A Multiscale Video Transformer Network}

\author{Mallika Garg\orcidlink{0000-0002-6056-6490} \and
Debashis Ghosh\orcidlink{0000-0001-5672-7645} \and
Pyari Mohan Pradhan\orcidlink{0000-0002-6070-5577}}

\authorrunning{Garg M. et al.}

\institute{Department of Electronics and Communication Engineering, \\ Indian Institute of Technology, Roorkee,  India\\
\email{mallika@ec.iitr.ac.in, debashis.ghosh@ece.iitr.ac.in, pyarimohan.pradhan@gmail.com}\\}

\maketitle
\begin{abstract}
In this paper, we introduce a novel Multiscale Video Transformer Network (MVTN)  for dynamic hand gesture recognition, since multiscale features can extract features with variable size, pose, and shape of hand which is a challenge in hand gesture recognition. The proposed model incorporates a multiscale feature hierarchy to capture diverse levels of detail and context within hand gestures which enhances the model's ability. This multiscale hierarchy is obtained by extracting different dimensions of attention in different transformer stages with initial stages to model high-resolution features and later stages to model low-resolution features. Our approach also leverages multimodal data, utilizing depth maps, infrared data, and surface normals along with RGB images from NVGesture and Briareo datasets. Experiments show that the proposed MVTN achieves state-of-the-art results with less computational complexity and parameters. The source code is available at https://github.com/mallikagarg/MVTN.

\keywords{Dynamic Gesture Recognition  \and Video Transformer Network  \and Multiscale Multi-head Attention}
\end{abstract}

\section{Introduction}
Although, vanilla transformer is able to outperform the state-of-the-art methods in various task like image classification~\cite{dosovitskiy2020image}, gesture recognition~\cite{d2020transformer}, object detection~\cite{zhou2022centerformer}, EEG classification~\cite{siddhad2024efficacy}, music generation~\cite{huang2018music}, semantic segmentation~\cite{li2022attention}, etc, it still pose some limitations.  These limitations are 
\begin{enumerate*}[label=(\alph*)]
\item  it outputs a feature map of a single scale since all the transformer stages have identical output dimensions which results in a low-resolution feature map, 
\item its usage for high-resolution tasks is limited since it extracts global features, and 
\item its computational and memory costs are relatively high, since the attention scales quadratically with the sequential data even for common input image size. \end{enumerate*}

\begin{figure}				
\centerline{\includegraphics[width=12.3cm,height=4.2cm]{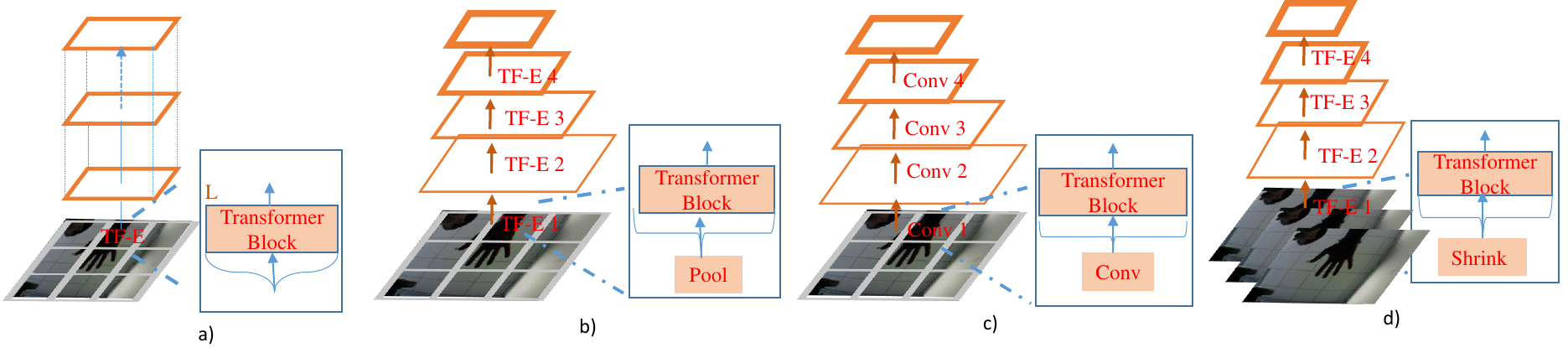}}
\caption{ Comparison of different transformer models, where TF-E is Transformer encoder. a) ViT~\cite{dosovitskiy2020image}, which has the same dimension of the attention for various stages of the transformer block (columnar structure).    Other variants of ViT have a hierarchical structure.  A pyramid of features is learned using:  b) a pooling operator, MviT~\cite{fan2021multiscale}, c) a convolution projection, CvT~\cite{wu2021cvt} and d) a linear projection, which is used in our model (MVTN) to progressively shrink the extracted features and take the advantage of scaling.}
\label{fig1}
\end{figure}

In the literature, many methods tackle these challenges in the transformer model by introducing pyramidal structures~\cite{heo2021rethinking, chu2021twins} and using linear attentions~\cite{choromanski2020rethinking, xiong2021nystromformer,wang2020linformer}.  The pyramid structure extracts multiple levels of features at different resolutions. Initially, transformers were introduced to consider convolution-free models as they rely on a self-attention mechanism rather than convolution operation. However, with time, to tackle the above-mentioned challenges in the transformers, pyramid structures were introduced using convolution layers with the attention mechanism~\cite{deshmukh2024textual, wu2021cvt, li2023uniformer, liu2021swin, gu2022multi}.  These pyramid-structured transformers maintain computational efficiency and reduce the dimension of the sequence length significantly. 

Later, researchers worked towards tackling the issues in transformers in a convolution-free environment by progressively shrinking the pyramid structure using a pooling operation. This operation enables it to capture both local and global contextual information efficiently~\cite{wang2021pyramid,wu2022p2t,fan2021multiscale, li2112improved,zheng2023potter, garg2023multiscaled}. The pyramid shrinking could either be done by scaling the feature map by employing a patch embedding layer or by pooling the sequence length. Pyramid structure models the low-resolution features in the initial layers and higher-level features in the deeper layers. These networks further reduce the computational cost as pooling layers do not have any learnable parameters. Pooling highlights the important features by selecting the strongest features within each region and capturing rich contextual information which maintains the performance of the transformer with less computation.

Keeping in view the aforementioned aspects and challenges of the transformer model, we propose a novel multiscale transformer network, MVTN, a network design for dynamic hand gesture recognition with less computation compared to the vanilla transformer. The intuition behind designing MVTN is to extract multiscale attention features from different stages of the transformer model. This enables the model to extract contextual information at different scales which helps to tackle the hand pose, size, and shape variations which is a significant challenge while designing robust hand gesture recognition models since the shape and size of the hand or fingers varies with signers.  In the proposed model, we linearly project the input dimension to the dimension of interest to design a pyramid hierarchy of transformer stages. This way, we create a pyramid of attentions at different stages in the proposed MVTN model which helps to learn multiscale attentions in the hand gesture, solving the problem of scale and size variance.   Fig.~\ref{fig1} shows a comparison of all the structures of the transformer models using the same attention throughout the network and using a hierarchical structure which is achieved by either using convolutional layers, pooling layers, or linearly projecting the input dimension to the dimension of interest.

We adapt the idea of a pyramid structure to reduce the computation of the transformer and at the same time learn multiscale attention with each transformer stage. By linearly projecting the input, we not only reduce the dimension and create a pyramid, but we also extract meaningful features that preserve the most relevant information. Since the linear layer has learnable parameters, it learns which features are most relevant for hand gesture recognition. Unlike pyramid pooling which aggregates features from local regions of the input feature, linear projection learns the features from the input in a way that the learned features are optimal for that particular task. Therefore, different from  P2T~\cite{wu2022p2t}, MViT~\cite{fan2021multiscale}, PVT~\cite{wang2021pyramid} and  MViTv2~\cite{li2112improved} which uses pyramid pooling and convolution, we use linear projection to reduce the dimension of attention at each stage and create a pyramid of attentions. 

Thus, the proposed MVTN has the following   major contributions:
\begin{enumerate}
\item We propose a pyramid hierarchy of multiscale attention to extract multiscale features from different stages of the transformer model. Multiscale attention captures hierarchical contextual information which helps the model to handle scale variations in the input hand gesture.

\item The pyramid structure is achieved by progressively shrinking the attention dimension using linear projection without the convolution, thus designing a convolution-less multiscale transformer for dynamic hand gesture recognition. An added advantage of the pyramid attention is a decrement in the computational cost.

\item Efficacy of the proposed MVTN model is accessed with NVGesture~\cite{molchanov2016online} and Briareo~\cite{manganaro2019hand} dataset with single and multi-modal inputs from active data sensors. Thus, our model achieves better performance compared to the existing state-of-the-art methods in hand gesture recognition.
\end{enumerate}

\section{Related Work}

\subsection{Transformer for Gesture Recognition}
Originally, transformers were designed for translation tasks in natural language processing (NLP)~\cite{vaswani2017attention}.  Initially, various deep learning methods were used for gesture recognition~\cite{mallika2022two, 9691523}. Later, transformers were used in gesture recognition~\cite{de-coster-etal-2020-sign}. In~\cite{d2020transformer}, multimodal input sequences can be processed for dynamic gesture recognition, which uses Video transformers~\cite{neimark2021video} as the base model. The attention in the spatial dimension is not enough to model the video sequence in dynamic hand gestures, so local and global multi-scale attention is proposed both locally and globally~\cite{s22062405}. The local attention extracts the information of the hand and the global attention learns the human-posture context. A combination of convolutions with self-attention is proposed in~\cite{s22062405}  for fusing spatial and temporal features for multimodal dynamic gesture recognition.  Recurrent 3D convolutional neural networks are also used in conjunction with transformer models for end-to-end learning for egocentric gesture recognition~\cite{Cao_2017_ICCV}. In~\cite{vaswani2017attention}, since the transformer must know the ordering of the sequence input, sinusoidal position embedding is added with the input. Initially, methods that used transformers for gesture recognition used sinusoidal positional encoding but later, a new  Gated Recurrent Unit (GRU)-based positioning scheme was incorporated into the Transformer networks~\cite{aloysius2021incorporating}.

\subsection{Convolutions in  Transformer}
High computational complexity is the major challenge in transformer models due to the large sequence length. To deal with it, researchers incorporated convolution in the transformers leveraging the convolution projections. This helps the model~\cite{wu2021cvt} to take advantage of convolution like scale, shift, dimension reduction with depth in layers, etc along with the merits of transformers like global context, attention mechanism, etc.   While Convolution neural networks (CNNs) are capable of reducing local redundancy,  they are not able to learn global features due to small receptive fields. Unifying  CNNs with transformers can leverage the benefits of both CNNs and transformers which can capture global and local features~\cite{li2023uniformer}.  Incorporating CNNs into transformers not only helps in less computations and learning local-global features but also helps in faster convergence~\cite{xiao2021early,yuan2021incorporating}.  LeVit~\cite{graham2021levit} introduces attention bias with convolution to integrate position information in the transformer for faster convergence. LeViT uses convolutional embedding instead of the patch-wise projection used in ViT.

With convolution, the spatial dimension decreases and the depth increases, which is beneficial for transformer architecture when convolution is used before attention block~\cite{yuan2021hrformer}.   Swin Transformer\cite{liu2021swin} were introduced with shifted windows using convolutions in transformers. The original self-attention block overlaps single low-resolution features at each stage of the transformer. Swin transformer creates a hierarchy of scaled non-overlapping windows which makes the model flexible to various scales. Axial attention is another attention mechanism that aligns multiple dimensions of the attention into the encoding and the decoding settings~\cite{ho2019axial}. Later, another hierarchical design that uses cross attention in horizontal and vertical strips to form the cross-shaped window was introduced in~\cite{dong2022cswin}. CSwin introduces Locally-enhanced Positional Encoding (LePE), which is a position encoding scheme that handles local positions. This provides CSwin with a strong power to model input into attention features with less computation.

\begin{figure}[tb]				
\centerline{\includegraphics[scale=.38]{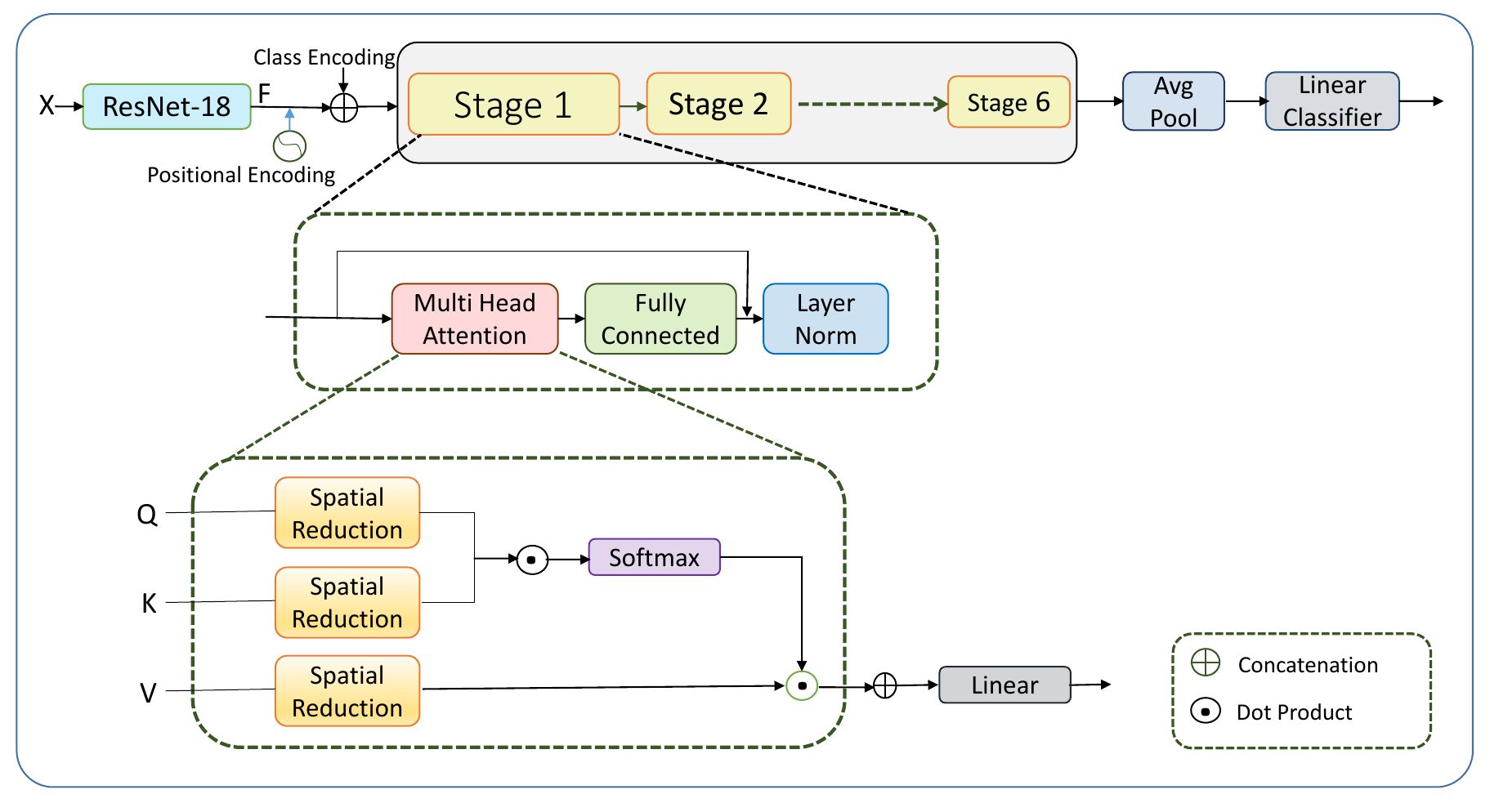}}
\caption{The overall architecture of the proposed Multiscale Video Transformer Network (MVTN) for dynamic hand gesture recognition.   Here, the size of the stage block of the transformer is kept varying to show the progressive reduction in the dimension of the attention vector with each stage.  Our proposed model captures multiscale contextual information of the hand gesture, which helps to tackle the major challenges of hand shape and size variations.}
\label{fig2}
\end{figure}
\subsection{Pooling in  Transformer}
In computer vision, pooling operation helps extract important or more powerful features. In the literature, pooling is used in CNNs for various tasks eg, semantic segmentation,  recognizing natural scenes, image classification, object detection, etc. Later, pyramid pooling~\cite{he2015spatial} has been widely applied to various computer vision tasks. Nowadays, pooling is used in transformers to learn powerful contextual representations and reduce the sequence length simultaneously. MViT~\cite{fan2021multiscale}  proposes a multiscale hierarchy of features for transformers using Multi Head Pooling Attention  (MHPA) which uses pooling attention.  The multi-scale stages expand the channel capacity while reducing the spatial resolution. An improvement over MViT is it uses decomposed relative positional embeddings and residual pooling connections with hybrid window mechanism~\cite{li2112improved}. Similarly, Pyramid Vision Transformer (PVT)\cite{wang2021pyramid}, is also a pyramid structure that is a pure transformer-based architecture used for object detection but it uses patch embedding for spatial reduction of the input. 

An advancement over PVT and MViT is proposed in P2T~\cite{wu2022p2t} which uses multiple pooling operations with different receptive fields and strides on the input feature. This helps  P2T to squeeze the input to me powerful contextual representations. A token pooling is proposed in PSViT~\cite{chen2021psvit} for scene understanding to aggregate information from multiple tokens in the transformer and attention sharing across different layers of the model. These two mechanisms help in the reduction of the computational cost and parameters of ViT. Since these pooling pyramid structures are lightweight and efficient, these can be used for real-time applications like POTTER~\cite{zheng2023potter} (Pooling Attention Transformer) which is designed for Human Mesh Recovery. 

\section{Multiscale Transformer}
\begin{figure}[tb]				
\centerline{\includegraphics[scale=.332]{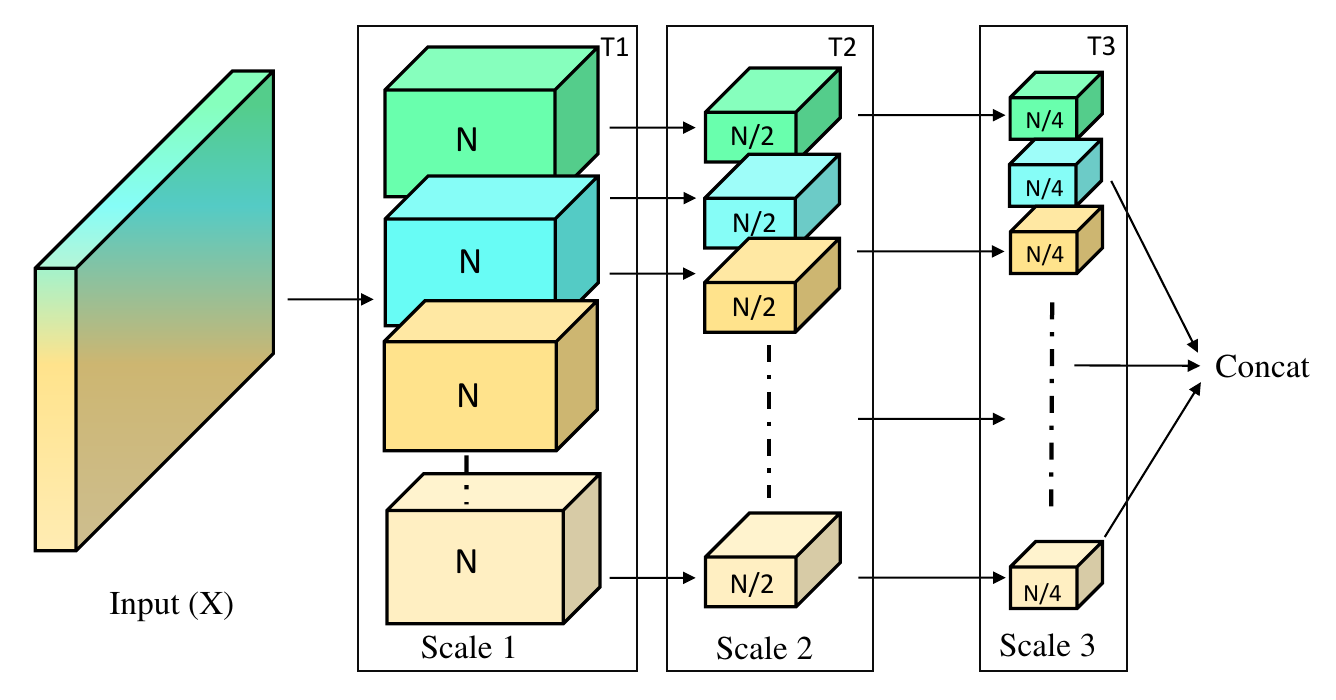}}
\caption{The proposed Multiscale Attention learns a pyramid hierarchy with 6 transformer stages (T1-T6), where $N = B\times T \times D$ is the dimension of attention in the first stage.  For convenience, pyramid scaling at each stage with a factor of  $N/2$ from the previous stage is shown only for three stages.}
\label{fig3}
\end{figure}

\subsection{Overview}\label{formats}
The proposed gesture recognition framework is designed to take a sequence of dynamic hand gestures as input and predict the class label of that particular video sequence. An overview of the proposed MVTN is shown in Figure~\ref{fig2}. Let X be the input dynamic gesture sequence such that it consists of $m$ frames, then it can be represented as $X = \{x_{1}, x_{2},..,x_{m}\}$. Here, $x_m$ denotes the frame at $m_{th}$ time instant and $X \in \mathbb{R}^{m \times w \times h \times c}$.

First, the feature map $f_t^m$ from the $m^{th}$ frame of the input gesture sequence is extracted using ResNet-18~\cite{he2016deep} of size $k$ which is then concatenated to get the feature map of the complete input hand gesture sequence. It can be denoted as a function $F $ over complete sequence $X$
such that,
\begin{equation}
\label{eqn:100}
F(X) = f_t^1 \oplus f_t^2 \oplus \dots \oplus   f_t^m 
\end{equation}
where $F: \mathbb{R}^{m \times w \times h \times c} \rightarrow \mathbb{R}^{m \times k}$ is a function of feature extractor.

The extracted features of the complete sequence are then encoded using spatial embedding. Position embedding is added to the encoded features followed by the addition of class token embedding~\cite{jeevan2022resource}. This encoded feature is then given as input to the proposed MVTN which acts as a temporal attention module. Before feeding We adapt the transformer to extract multiscale attention features so that the model can learn variations in hand shape and size efficiently. Sine-cosine position embedding~\cite{vaswani2017attention}  is added to each frame before feeding it to the transformer block since the model should know the ordering of the sequence. Finally, the classification head is used to get the probability distribution over $n$ classes. 

\subsection{Multiscale Attention Pyramid}\label{4}
In the proposed Multiscale Attention, we linearly project the Query (Q), Key (K), and Value (V) vectors to form the pyramid structure with the transformer stages. This means that at every stage, the model learns a different dimension of attention features. Fig~\ref{fig3} shows the scaling attention features at every stage by a factor of 2 from the previous stage. So, if the previous stage has $N$ features,  it is scaled by $N/2$, where $N = B\times T \times D$ for the first stage.  Since there are 6 stages, it becomes $N$, $N/2$, $N/4$, $N/8$, $N/16$ and $N/32$.  Like other pyramid transformers~\cite{wang2021pyramid,wu2022p2t}, we also progressively shrink the attention dimensions to learn a hierarchy of multiscale features. For convenience, we represent $B\times T$ as $L$ in the rest of the paper, where $B$ is the batch size, and $T$ is the number of frames.

The input to the proposed model is a sequence of frames, $X \in \mathbb{R}^{ L \times D}$, where $D$ is the dimension of features extracted from ResNet-18 which is the same as the input of the first stage of the transformer block, $d_{model} = 512$. The proposed MVTN is a fully convolution-free model, as we have used linear projections to reduce the dimension of the input vector and not the convolutional layer. 

\begin{figure}[tb]				
\centerline{\includegraphics[scale=.48]{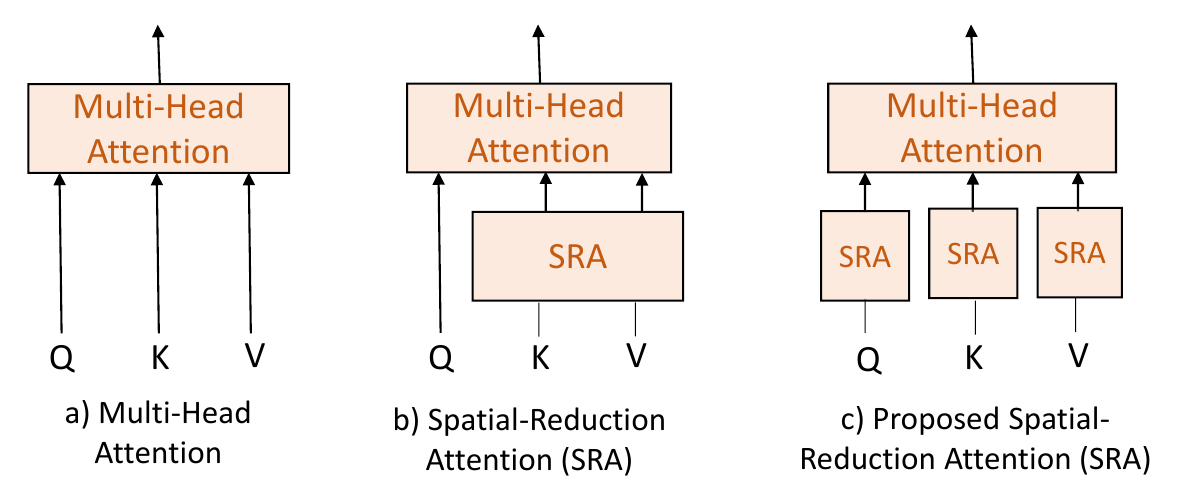}}
\caption{ a) Multi-head attention (MHA)~\cite{vaswani2017attention}, b) Spatial-Reduction Attention (SRA)~\cite{wang2021pyramid}, c) Proposed spatial-reduction attention. We spatially reduce the query (Q), Key (K), and Value (V) using linear projection while in b) only the key and value are reduced.   This spatial reduction helps to incorporate a multiscale pyramid structure along with the reduction in the computational cost of the model.}
\label{fig4}
\end{figure}

\subsection{Spatial-Reduction Attention}\label{5}
We propose a Spatial-Reduction Attention block that linearly reduces the dimension of the attention as shown in Fig.~\ref{fig4}. Similar to Multi-head attention (MHA)~\cite{vaswani2017attention} and Spatial-Reduction Attention (SRA)~\cite{wang2021pyramid}, our SRA block also receives three vectors as input, Query (Q), Key (K) and Value (V).  Different from MHA which has a columnar structure,  PVT and MVTN use spatial reduction blocks to project the input vectors to lesser dimensional sub-space. Also, our proposed spatial reduction block employs reducing dimensions of query vector along with key and value, unlike PVT which only reduces key and value dimension.  This further helps in the reduction of computational cost compared to PVT. Downscaling the query helps the model to better generalization by encouraging the model to focus on more informative features while ignoring irrelevant ones. 

Mathematically,  if  $X$  is the input to the proposed model, it is linearly projected to scale query, key, and value vectors, in a way that $\textbf{Q} \in \mathbb{R}^{L\times D}$, $\textbf{K} \in \mathbb{R}^{L\times D}$ and $\textbf{V}  \in \mathbb{R}^{L\times D}$, respectively. The three attention vectors of a stage can be represented as
\begin{equation}
\label{eqn:5}
\textbf{Q}_{jM/2} = XW^\textbf{Q}_{jM/2}, ~\textbf{K}_{jM/2} = XW^\textbf{K}_{jM/2}, ~ \textbf{V}_{jM/2} = XW^\textbf{V}_{jM/2},
\end{equation} 
where $j \in (1, 6) $ for 6 stages and $M$ is the dimension of the attention tensor of the previous stage.  

\subsection{Multi-Modal Late Fusion}\label{9}
Recently, multimodal methods have attained significant attraction and it is used in multiple applications. Since, the dynamic hand gesture datasets, NVGesture~\cite{molchanov2016online} and Briareo~\cite{manganaro2019hand} acquire the data from RGB-G sensors, they have provided RGB images, depth images, and infrared images as inputs. These inputs are independently given to the model to predict the probability score of each class. Following~\cite{d2020transformer}, we have adopted a late fusion technique that aggregates the predictions or scores obtained from each modality independently. Late fusion is commonly used when the input is taken from multiple sensors as it can yield comparable results or improve overall performance by effectively leveraging complementary information from different sources. We have  performed averaging over all the probabilities obtained from single modal inputs to generate the final prediction or decision of a combination of different numbers of inputs, which is given as

\begin{equation}
\label{eqn:6}	
y = \arg\max_j \sum_{i}^n P(\omega_j|x_i),
\end{equation}
where $n$ is the number of modalities over which the results are to be aggregated, and  $P(\omega_j|x_i)$ is the probability distribution of the  $i^{th}$ frames of a given input, which belongs to class $\omega_j$.

\begin{table}[tb]
\caption{Details of attention dimension in the proposed MVTN model.}
\centering
\begin{tabular}{c|c|c|c}
\hline
\multirow{2}{*}{Stages}& \multicolumn{3}{c}{Attention Input Tensor size}  \\  \cline{2-4}
& $P_{dim}$ &  $P_{dim_{rev}}$ & $P_{dim_+}$ \\
\hline
$Stage_1$& $B\times T\times D/32$& $B\times T\times D$ & $B\times T\times (D/2+D/64)$\\
$Stage_2$& $B\times T\times D/16$& $B\times T\times D/2$& $B\times T\times (D/2+D/32)$\\
$Stage_3$& $B\times T\times D/8$ &$B\times T\times D/4$& $B\times T\times (D/2+D/16)$\\
$Stage_4$&  $B\times T\times D/4$ & $B\times T\times D/8$& $B\times T\times (D/2+D/8)$\\
$Stage_5$& $B\times T\times D/2$ &  $B\times T\times D/16$& $B\times T\times (D/2+D/4)$\\
$Stage_6$&  $B\times T\times D$ & $B\times T\times D/32$& $B\times T\times (D/2+D/2)$\\
\hline	
\end{tabular}
\label{tab6}
\end{table}

\section{Experiments and Discussion}\label{6}
We have experimented on the two publicly available dynamic hand gesture datasets: NVGesture and Briareo. We have tested our proposed MVTN model with single as well as multi-modal inputs.  We have also compared the proposed model with the other methods and discussed the results obtained. Further, we have also performed the ablation analysis with different model settings.  We have experimented on different attention input dimensions to form the pyramid structure w.r.t to the stage of the transformer as given in Table~\ref{tab6}. First, we experimented with  $P_{dim}$, which projects the input tensor to fewer dimensions in the initial stages, while progressively increasing the dimension of the attention with each stage forming a pyramid of transformer stages. This way the model learns low-resolution attention features followed by high-resolution.

MViT, PVT, P2T, POTTER, and other pyramid-structured transformer-based models in the literature, learn high-resolution features while progressively decreasing the resolution. So, along with $P_{dim}$, we also experimented with the same pyramid pattern as in ~\cite{ fan2021multiscale, wang2021pyramid,wu2022p2t,zheng2023potter}, we name it as $P_{dim_{rev}}$. Since, we employ the same configuration as in the base model~\cite{vaswani2017attention} having 6 transformer stages and output dimension $D= d_{model}$ =512, which is linearly projected in 8 heads of the 64-dimensional vector. Thus, creating a pyramid structure with $N/2$ reduction at each stage leaves the first and last stage of $P_{dim}$ and $P_{dim_{rev}}$ with a 16-dimensional vector, respectively. Further, dividing this vector into 8 heads will yield a vector of just 2-dimension. To our intuition learning such a small vector will not learn enough information. So, we experimented with one more attention pyramid structure,  $P_{dim_+}$.  In this, we set the dimension of each stage such that the last stage of the pyramid has a maximum $D/2 + D/2 = D$ dimension and each stage has at least $D/2$ dimension. In this way, the dimension of the first stage is  $(D/2+D/64)$, which progressively increases till the last stage.

\subsection{Datasets}\label{77}
\textbf{NVGesture:}
NVGesture~\cite{molchanov2016online} is the largest dataset with 25 dynamic gesture classes. Each gesture is recorded with multiple sensors and views, front and top view of the driver in the car simulator under different lighting conditions. A total of 1532 dynamic hand gestures were acquired by a diverse group of 20 subjects with 3 modalities (RGB, IR, and depth). Each gesture is performed 3 times.  There are 5 streams of data present in the dataset: color, depth, color mapped on depth, IR left, and IR right. Along with this, optical flow from the RGB images is computed using~\cite{farneback2003two}, which is made publicly available along with the dataset.
\begin{table}[ht]
\caption{Results for different modalities on NVGesture~\cite{molchanov2016online}  and Briareo~\cite{manganaro2019hand} dataset. \# is the number of input modalities used.}
\centering
\begin{tabular}{c|ccccc|cc|cc}
\hline
\multirow{3}{*}{\#}&\multicolumn{5}{c|}{Input data}& \multicolumn{4}{c}{Accuracy} \\ \cline{2-10}

&\multirow{3}{*}{Color} &\multirow{3}{*}{ Depth}& \multirow{3}{*}{IR} & \multirow{3}{*}{Normals} & \multirow{3}{*}{Optical flow} & \multicolumn{2}{c|}{NVGesture} & \multicolumn{2}{c}{Briareo}\\  \cline{7-10}

&&&&&&Transformer & \multirow{2}{*}{MVTN} & Transformer & \multirow{2}{*}{MVTN}\\
&&&&&&~\cite{d2020transformer}&&~\cite{d2020transformer}&\\
\hline

\multirow{4}{*}{1} 
&\checkmark&&&&&    76.50\%&77.50\%& 90.60\%&97.69\%  \\
&&\checkmark&&&&    83.00\%&\textbf{85.21}\%& 92.40\%&97.92\%\\
&&&\checkmark&&&    64.70\%&70.42\%& 95.10\%&96.30\%\\
&&&&\checkmark&&    82.40\%&83.75\%& 95.80\%&\textbf{98.26}\%\\ \hline 

\multirow{9}{*}{2}
&\checkmark&\checkmark&&&&  84.60\% &83.61\%&94.10\%&97.57\%\\
&\checkmark&&\checkmark&&&  79.00\%&77.59\%&95.50\%&97.22\%\\
&&\checkmark&\checkmark&&&  81.70\%&83.61\%&95.10\%&97.57\%\\
&\checkmark&&&\checkmark&&  84.60\%&84.44\%&96.50\%&\textbf{97.92}\%\\
&&\checkmark&&\checkmark&&  87.30\%&\textbf{85.68}\%&96.20\%&96.56\%\\
&&&\checkmark&\checkmark&&  83.60\%&83.82\%&97.20\%&97.57\%\\ 
&\checkmark&& &&\checkmark& 72.00\% &83.61\%&-&96.57\%\\
&&\checkmark&&&\checkmark&  -&83.61\%&-&97.91\%\\
&&&\checkmark&&\checkmark&   -&83.60\%&-&97.57\%\\ \hline

\multirow{4}{*}{3} 
&\checkmark&\checkmark&\checkmark&&& 85.30\% &\textbf{87.80}\%&95.10\%&96.88\%\\
&\checkmark&\checkmark&&\checkmark&& 86.10\%&85.68\%&95.80\%&\textbf{98.61}\%\\
&\checkmark&&\checkmark&\checkmark&& 85.30\%&83.61\%&96.90\%&96.88\%\\
&&\checkmark&\checkmark&\checkmark&& 87.10\%&83.60\%&97.20\%&\textbf{98.6}1\%\\ 
\hline

\multirow{4}{*}{4}
&\checkmark&\checkmark&\checkmark&\checkmark&&  87.60\%&86.30\%&96.20\%&97.22\%\\
&\checkmark&\checkmark&&\checkmark&\checkmark&  -&84.85\%&-&\textbf{98.61}\%\\ 
&\checkmark&&\checkmark&\checkmark&\checkmark&  -&\textbf{87.31}\%&-&97.92\%\\ 
&&\checkmark&\checkmark&\checkmark&\checkmark&  -&85.06\%&-&\textbf{98.61}\%\\

\hline
5&\checkmark&\checkmark &\checkmark&\checkmark&\checkmark& -&84.85\%&-&97.92\%\\
\hline

\end{tabular}
\label{tab1}
\end{table}

\textbf{Briareo:}
Briareo dataset~\cite{manganaro2019hand} contains dynamic hand gestures collected using a RGB camera, a depth sensor, \textit{Pico Flex} and an infrared stereo camera, the Leap Motion under natural lightening conditions, as a result of which the images were dark with low contrast. RGB cameras can collect videos at 30 frames per sec. The depth sensor can acquire a depth map of spatial resolution $224 \times  171$. Finally, the Leap Motion has 2 infrared cameras with $ 640 \times 240$ and $200 \times 200$ resolution.  All the 3 devices are synchronized so that the data acquired at a certain instant depict the same time instant in that particular sequence. 12 dynamic gestures were performed by 40 subjects out of which  33 are males and 7 are females.  Each gesture is performed 3 times by every subject. Thus a total of 120 ($40 \times 3$) sequences of each gesture is collected of at least 40 frames.

\begin{table}[ht!]
\caption{Comparison results for single modality on NVGesture dataset~\cite{molchanov2016online} }
\centering
\begin{tabular}{ccc}
\hline
Input modality&Method& Accuracy \\ 
\hline
\multirow{11}{*}{Color} 
&Res3ATN~\cite{dhingra2019res3atn}& 62.70\%\\
&C3D~\cite{tran2015learning}&69.30\%\\ 
&R3D-CNN~\cite{molchanov2016online}& 74.10\%\\
&GPM~\cite{fan2021multi}&75.90\% \\
&PreRNN~\cite{yang2018making}&76.50\% \\
&GestFormer~\cite{Garg_2024_CVPR}&75.41\% \\ 
&Transformer~\cite{d2020transformer}& 76.50\%\\
&I3D~\cite{wang2016robust}&78.40\%\\

&Human~\cite{molchanov2016online}&88.40\% \\
&\textbf{MVTN}& \textbf{77.50\%}\\
\hline

\multirow{9}{*}{Depth}
&R3D-CNN~\cite{molchanov2016online}& 80.30\%\\
&I3D~\cite{wang2016robust}&82.30\%\\
&GestFormer~\cite{Garg_2024_CVPR}&80.21\% \\ 
&Transformer~\cite{d2020transformer}&83.00\%\\
&ResNeXt-101~\cite{kopuklu2019real}& 83.82\%\\
&PreRNN~\cite{yang2018making}&84.40\% \\
&MTUT~\cite{abavisani2019improving}&84.85\%\\
&\textbf{MVTN}& \textbf{85.21\%}\\ \hline	

\multirow{7}{*}{Optical flow}
&Temp. st. CNN~\cite{simonyan2014two} & 68.00\%\\
&Transformer~\cite{d2020transformer}& 72.61\%\\
&GestFormer~\cite{Garg_2024_CVPR}&81.66\% \\ 
&iDT-MBH~\cite{vadisaction} & 76.80\%\\
&R3D-CNN~\cite{molchanov2016online} & 77.80\%\\
&I3D~\cite{wang2016robust} & 83.40\%\\
&\textbf{MVTN}&\textbf{72.50\%}\\  \hline

\multirow{2}{*}{Normals}&Transformer~\cite{d2020transformer} &82.40\% \\
&GestFormer~\cite{Garg_2024_CVPR}&81.66\% \\ 
&\textbf{MVTN}& \textbf{83.75\%}\\ \hline

\multirow{3}{*}{Infrared}&R3D-CNN~\cite{molchanov2016online}& 63.50\%\\
&GestFormer~\cite{Garg_2024_CVPR}&63.54\% \\ 
& Transformer~\cite{d2020transformer}&  64.70\% \\
&\textbf{MVTN}&\textbf{70.42\%}\\ \hline

\end{tabular}
\label{tab2}
\end{table}

\subsection{Implementation Details}\label{7}
We implemented the proposed MVTN model on  Nvidia GeForce GTX 1080 Ti 12 GB GPU, CUDA 10.2  with cuDNN 8.1.1 and Torch 1.7.1.  The model is trained using the Adam optimizer on a sequence length of 40 frames with a batch size of 8 with a learning rate of $1e^{-4}$ and a weight decay at $50^{th}$ and $75^{th}$ epoch over the categorical cross-entropy loss. We follow~\cite{d2020transformer} and cropped the image to $224 \times224 $ to extract features from the ResNet-18 model pre-trained on the ImageNet dataset~\cite{deng2009imagenet}. We employ various data augmentation techniques such as rotation, scaling, and cropping to prevent overfitting. We have separately trained models for each modality and performed decision-level fusion using late fusion. 
\subsection{Results and Discussion}
\textbf{NVGesture:} 
Following~\cite{d2020transformer}, we have performed experiments with single and multi-modality. We trained the model separately for each modality and then used late fusion as discussed in Section~\ref{9}. The results obtained on the NVGesture dataset for different modalities on the proposed model and the transformer~\cite{d2020transformer}  baseline are reported in Table~\ref{tab1}. Here, we have reported the best results out of the 3 pyramid dimension settings for each modality.  From the table, we observe that MVTN achieves far better results than the traditional transformer used for gesture recognition~\cite{d2020transformer} on a single modality.

 \begin{table}[ht]
\caption{Comparison results for multi-modalities on NVGestures dataset~\cite{molchanov2016online}. }
\centering
\begin{tabular}{ccc}
\hline
Method & Input modality & Accuracy \\ 
\hline

iDT ~\cite{vadisaction}& color + flow & 73.00\% \\
\hline
R3D-CNN~\cite{molchanov2016online} & depth + flow & 82.40\%\\
R3D-CNN~\cite{molchanov2016online} & all & 83.80\%\\
\hline
MSD-2DCNN~\cite{fan2021multi}&color+depth&84.00\% \\

\hline
8-MFFs-3f1c\cite{kopuklu2018motion}&color + flow& 84.70\%\\
\hline
STSNN~\cite{zhang2020dynamic}&color+flow& 85.13\%\\
\hline
PreRNN~\cite{yang2018making}& color + depth&85.00\% \\
\hline

I3D~\cite{wang2016robust}& color + flow &84.40\%\\
I3D~\cite{wang2016robust}& color + depth + flow &85.70\%\\

\hline
{GestFormer~\cite{Garg_2024_CVPR}}&depth + color + ir +normal&{85.62\%}\\ 
{GestFormer~\cite{Garg_2024_CVPR}}&depth + color + ir + normal + op&{85.85\%}\\

\hline	
GPM~\cite{fan2021multi}& color + depth&86.10\% \\
\hline
MTUT\textsubscript{RGB-D}~\cite{abavisani2019improving}& color + depth& 85.50\%\\
MTUT\textsubscript{RGB-D+flow}~\cite{abavisani2019improving}& color + depth& 86.10\%\\
MTUT\textsubscript{RGB-D+flow}~\cite{abavisani2019improving}& color + depth + flow& 86.90\%\\
\hline

Transformer~\cite{d2020transformer}& depth + normals &87.30\%\\
Transformer~\cite{d2020transformer}& color + depth + normals+ir& 87.60\%\\
\hline

NAS2~\cite{yu2021searching}& color + depth&86.93\% \\
NAS1+NAS2~\cite{yu2021searching} &color + depth&88.38\% \\
\hline
\textbf{MVTN}&\textbf{depth + normals}&\textbf{85.64\%}\\	
\textbf{MVTN}&\textbf{depth + color+ ir }&\textbf{87.80\%}\\ 
\hline	
\end{tabular}
\label{tab3}
\end{table}

Our model shows better results with approx. 8.84\% of increase in the accuracy when evaluated on infrared.  Though the results on depth images show an increment of  2.66\%, we obtain the best result on depth in a single modality with an accuracy of 85.21\%. We also observe that normals also have nearly the best accuracy which shows that normals can be a good representation of hand gesture which is driven from depth input. To further improve the accuracy, we performed late fusion and observed that the accuracy increased up to 85.68\% with double modality when depth and normal probabilities were fused. Further, we observe that the accuracy shows improvement when 3 modes of inputs are fused.  We obtain the best accuracy of 87.80\% on the NVGesture dataset with triple modality (color, depth, and ir). Further, we have experimented with 4 and 5 modes of input, but the best results were obtained for only 3 modes.

We further compare the obtained results with the other methods on a single modality in Table~\ref{tab2} and observe that our model can obtain state-of-the-art results. Similarly, we compare the results on multimodal inputs in Table~\ref{tab3} and conclude that MVTN has better results with only 3 modalities as compared to ~\cite{d2020transformer} which has even lesser results with 4 modalities.

\begin{table}[tb]
\caption{Comparison of the results obtained for different modalities on Briareo dataset~\cite{manganaro2019hand}.}
\centering
\begin{tabular}{ccc}
\hline
Method& Inputs & Accuracy  \\ 
\hline
C3D-HG~\cite{manganaro2019hand}& depth& 76.00\%\\
C3D-HG~\cite{manganaro2019hand}& ir& 87.50\%\\

LSTM-HG~\cite{manganaro2019hand}&3D joint features &94.40\%\\
\hline
NUI-CNN~\cite{d2020multimodal}& depth + ir& 92.00\%\\
NUI-CNN~\cite{d2020multimodal}& color + depth + ir& 90.90\%\\

\hline
Transformer~\cite{d2020transformer}& normals& 95.80\%\\
Transformer~\cite{d2020transformer}& depth + normals &96.20\%\\
Transformer~\cite{d2020transformer}&ir + normals &97.20\%\\
\hline	
GestFormer~\cite{Garg_2024_CVPR} & ir&98.13\%\\
GestFormer~\cite{Garg_2024_CVPR} & ir + normals &97.57\%\\
\hline
\textbf{MVTN} & \textbf{normals}&\textbf{98.26} \%\\
\textbf{MVTN} & \textbf{color + depth + normals} &\textbf{98.61}\%\\
\textbf{MVTN} &\textbf{depth + ir + normals} &\textbf{98.61}\%\\
\hline
\end{tabular}
\label{tab5}
\end{table}

\begin{table}[tb]
\caption{Ablation with different dimensions of pyramid structure and embeddings on NVGesture~\cite{molchanov2016online} and Briareo dataset~\cite{manganaro2019hand} dataset (top and bottom, respectively). }
\centering
\begin{tabular}{c|ccc|ccc}
\hline
\multirow{2}{*}{Input modality}&\multicolumn{3}{c|}{w/o embeddings} &\multicolumn{3}{c}{w embeddings} \\ \cline{2-7}
&$P_{dim}$ &$P_{dim_{rev}}$&$P_{dim_+}$&$P_{dim}$ & $P_{dim_{rev}}$&$P_{dim_+}$\\ 
\hline
Color&77.08\% &\textbf{77.50}\% &   75.63\% &   75.83\%&   75.63\% &   65.83\% \\ 
Depth& 83.75\%   & 81.88\% &  \textbf{85.21}\%&  82.29\%&  82.50\%  &  83.33\%\\
Normal & \textbf{85.27}\% & 83.33\% & 82.92\%   & 76.88 \%&  83.75\%& 75.42 \% \\ 
IR  & \textbf{70.53}\% &68.59\%  & 68.12\%  &  60.00\%  & 67.29 \% &  58.96\%  \\ 
Color+ Optical flow & 72.50\%& 74.38\%& \textbf{75.20}\%& 74.83\%& 71.25\%& 72.92\% \\
Depth + Optical flow & \textbf{82.99}\%& 81.46 \%& 82.08\%& 82.08\%& 81.67\% &81.47\%\\ 
IR + Optical flow &\textbf{72.19}\%& 72.08\%& 69.58\%& 72.04\%& 71.67\%& 69.58\% \\ 

\hline \hline
Color&                  96.53\%  & \textbf{97.69}\% & 97.22\%& 96.53\% & 96.88\% & 97.22 \%\\
Depth&                  95.14\%  & 96.29\% & \textbf{97.92}\%& 97.22\% & 95.83\% & 96.29\%\\
Normal &                96.53\%  & 96.29\% & 96.29\%& 97.92\% & \textbf{98.26}\% & 96.30 \%\\ 
IR  &                   95.49 \% & \textbf{96.30}\% & 95.83\%& 94.79\% & 94.91\% &  95.83\%\\
Color+ Optical flow &   94.44\%  & 93.98\% & 94.91\% & \textbf{95.49}\%&95.37\% & 94.44\%\\
Depth + Optical flow &  94.37\%  & 92.98\% & 94.34\%& 94.44\% & 94.21\% & \textbf{94.91}\%\\
IR + Optical flow    &   95.21\% & \textbf{95.37}\% & 93.98\%&94.79 \% & 94.44\% & 94.91\%\\
\hline
\end{tabular}
\label{tab22}
\end{table}	

\textbf{Briareo:} 
Table~\ref{tab1} also presents the results for the Briareo dataset for single and multimodal inputs compared with~\cite{d2020transformer}. We observe that MVTN can obtain better results on all the modalities individually as well as the corresponding result of each experiment. We further observe an increment of a maximum of 7.82\% when experimented on color images with an accuracy of 97.67\% and the best accuracy on single input is 98.26\% which is obtained on normals. Combining modalities doesn't provide an improvement in the performance of MVTN but further increases the modality, and increases the accuracy to 98.61\% when color, depth, and surface normals are fused or depth, infrared, and normals are fused.  

Further, we compare the results obtained with different methods on single as well as multimodal inputs in Table~\ref{tab5} and observe that MVTN outperforms the other methods with 98.61\% accuracy. Finally, we can also conclude from the results that  MVTN can achieve better results on even a single modality by a clear margin compared to the state-of-the-art methods with single as well as multimodal inputs. Further, multimodal inputs add to the better performance of MVTN.

\subsection{Ablation}
We experimented with 3 different dimensions of the pyramid structure $P_{dim}$, $P_{dim_{rev}}$ and $P_{dim_+}$ as mentioned in Section~\ref{6}. We have also performed experiments with spatial embedding and class token embedding as shown in Table~\ref{tab22}. From the results, we conclude that for the NVGesture dataset, MVTN outperforms without embeddings, since the best results of all the 6 experiments (3 pyramid dimension $\times$ 2 for embeddings (w/o or w)) for each modality are obtained when no embedding is used. Further, adding the embeddings degrades the performance. However, this is not the case with the Briareo dataset. Best results for normal, color with optical flow and depth with optical flow are obtained when embedding is used in the MVTN for gesture recognition.  We have also compared the numbers of learnable parameters of our model with other models and traditional transformer models as shown in Table~\ref{tab24}, which shows that less number of parameters are needed to train MVTN for dynamic hand gesture recognition tasks.

\begin{table}[tb]
\caption{Comparison in terms of the number of parameters (M)). }
\centering
\begin{tabular}{c|c|c}
\hline
Methods&Params & MACs (G)\\  \hline
R3D-CNN~\cite{molchanov2016online} & 38.00&-\\
C3D-HG~\cite{manganaro2019hand} & 26.70&- \\
Transformer\cite{d2020transformer} & 24.30&62.92\\
GestFormer~\cite{Garg_2024_CVPR}&24.08&60.40\\
 MVTN, $P_{dim}$&\textbf{19.55}& \textbf{60.22}\\
 MVTN,$P_{dim_{rev}}$&\textbf{19.55}&\textbf{60.22}\\
 MVTN,  $P_{dim_+}$ &\textbf{21.67}&\textbf{60.31}\\
\hline

\end{tabular}
\label{tab24}
\end{table}	

\section{Conclusion}
We proposed a novel Multiscale Video Transformer Network (MVTN)  for dynamic hand gesture recognition that learns the multiscale feature at different stages of the transformer. This helps to tackle the problem of hand shape and size variation by extracting contextual information at different levels in a hierarchical manner which helps to reduce the computational cost.   Evaluating the proposed model on NVGesture and Briareo datasets shows that our model is better than the traditional transformer model. From the extensive experiments, we can conclude that our MVTN model is so efficient that it can outperform other methods with single and multimodal input. Specifically, on the Briareo dataset, the results are better with single inputs compared to multimodal inputs in other methods. This concludes that our single modality outperforms other method's multimodal outputs. This also helps in the reduction of model complexity and parameters.
\bibliographystyle{splncs04}
\bibliography{main}
\end{document}